\documentclass[letterpaper]{article}
\usepackage{acl}
\usepackage{latexsym}
\usepackage{times}
\usepackage{helvet}
\usepackage{courier}
\usepackage{microtype}

\usepackage[T1]{fontenc}
\usepackage[utf8]{inputenc}

\usepackage{graphicx}
\usepackage{booktabs}
\usepackage{amsmath}
\usepackage{natbib}
\usepackage{xcolor}
\usepackage{dirtytalk}

\newcommand{\ye}[1]{}
\newcommand{\uvp}[1]{}
\newcommand{\todot}[1]{}

\newcommand{\divrit}{\textsc{DiVRit}}
\newcommand{\hebrewbert}{\textsc{AlephBertGimmel-small}}
\newcommand{\modelname}{\textsc{Hebrew-PIXEL}}
\DeclareMathOperator{\freq}{freq}

\title{Hebrew Diacritics Restoration using Visual Representation}

\author{
  Yair Elboher \and Yuval Pinter \\
  Faculty of Computer and Information Science \\
  Ben-Gurion University of the Negev \\
  Be'er Sheva, Israel \\
  \texttt{yairel@bgu.ac.il, uvp@cs.bgu.ac.il}
}

\begin{document}

\maketitle
\begin{abstract}
Diacritics restoration in Hebrew is a fundamental task for ensuring accurate word pronunciation and disambiguating textual meaning.
Despite the language's high degree of ambiguity when unvocalized, recent machine learning approaches have significantly advanced performance on this task.

In this work, we present \divrit{}, a novel system for Hebrew diacritization that frames the task as a zero-shot classification problem.
Our approach operates at the word level, selecting the most appropriate diacritization pattern for each undiacritized word from a dynamically generated candidate set, conditioned on the surrounding textual context.
A key innovation of \divrit{} is its use of a Hebrew Visual Language Model to process diacritized candidates as images, allowing diacritic information to be embedded directly within their vector representations while the surrounding context remains tokenization-based.

Through a comprehensive evaluation across various configurations, we demonstrate that the system effectively performs diacritization without relying on complex, explicit linguistic analysis.
Notably, in an ``oracle'' setting where the correct diacritized form is guaranteed to be among the provided candidates, \divrit{} achieves a high level of accuracy.
Furthermore, strategic architectural enhancements and optimized training methodologies yield significant improvements in the system's overall generalization capabilities.
These findings highlight the promising potential of visual representations for accurate and automated Hebrew diacritization.
\end{abstract}

\section{Introduction}
\label{sec:introduction}
Diacritics play a vital role in many writing systems, indicating pronunciation and distinguishing between different meanings of words~\cite{N_plava_2021,gorman-pinter-2025-dont}.
In Semitic scripts, which primarily represent consonants, the absence of these marks leads to many words having multiple valid interpretations depending on the context, and introduces significant lexical ambiguity~\cite{elgamal-etal-2024-arabic}.
By providing essential grammatical information, these marks, known as \textit{niqqud} in Hebrew, resolve this ambiguity, ensuring a single, intended meaning.
For example, the Hebrew consonantal string ``\textit{mlk}'' can be interpreted as ``king'' (\textit{melekh}), ``reigned'' (\textit{malakh}), and more, depending on context.
Despite their usefulness, diacritics are typically omitted in modern Hebrew writing, relying on readers' familiarity to infer the correct interpretation.
This contrasts with the linguistic complexity involved in restoring the correct diacritization, which demands both syntactic and semantic understanding.

\begin{figure}[t]
    \centering
    \includegraphics[width=\linewidth]{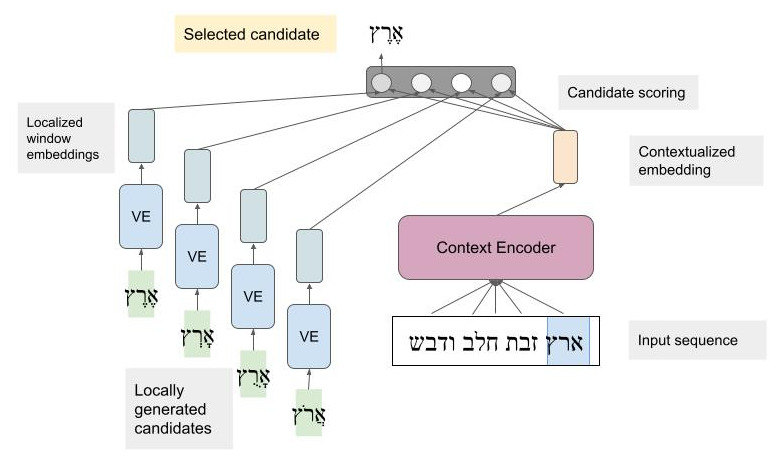}
    \caption{Architecture of the diacritizer.
    A context encoder (right) captures contextual information, while a visual candidate encoder (left) processes an aligned portion of each candidate from the candidate generator.
    The resulting embeddings are then compared in a scoring layer.}
    \label{fig:architecture}
\end{figure}

Automatic diacritization poses a significant challenge, especially for morphologically-rich languages like Hebrew.
Accurate automatic diacritization is important for both practical applications and theoretical advancements in language technology~\cite{chen-etal-2024-interplay}, contributing to improved accessibility in digital text and enhanced human-computer interaction.
However, despite a well-defined set of Hebrew diacritization rules, their application to text is non-trivial, requiring the extraction of information such as gender, tense, and part of speech~\cite{tsarfaty-etal-2019-whats}.
This complexity makes purely rule-based approaches impractical, necessitating alternative solutions, particularly machine learning-based approaches like neural networks.
Over the years, neural network-based Hebrew diacritization has progressed from hybrid to data-driven systems.
Dicta's Nakdan~\cite{shmidman-etal-2020-nakdan}, a system combining deep learning with linguistic rules, offers high accuracy but limited generalizability.
Conversely, Nakdimon~\cite{gershuni-pinter-2022-restoring} and MenakBERT~\cite{cohen-etal-2024-menakbert} leverage purely data-driven, character-level approaches, using a Bi-LSTM and a transformer pretrained on Hebrew corpora, respectively, achieving competitive performance with simplified, adaptable pipelines.

While these systems work at the character level, predicting each letter's niqqud symbols at a time, Dicta's approach of word-level analysis clearly works at a better-fitting granularity, as Hebrew morphology is mostly governed by word-level templates which operate beyond immediate and/or unidirectional, character contexts.
To this end, we propose a purely data-driven reframing of the Hebrew diacritization task as a zero-shot classification problem at the word level.
This entails generating a set of diacritization candidates for each word, with each candidate acting as a distinct class within this framework, calling for a zero-shot learning approach~\cite[ZSL;][]{8413121}.
The diacritization process then becomes the selection of the most contextually suitable class based on learned discriminative representations.
Unlike character-level classification methods, this strategy requires representations that effectively capture the subtle distinctions between diacritized word forms, thereby necessitating the use of a language model capable of processing diacritized text.

In this research we present \textbf{\divrit{}}, a \textbf{Di}acritizer with \textbf{V}isual \textbf{R}epresentation for Hebrew (I\textbf{VRIT}) text.
We explore the use of a vision transformer (ViT) architecture~\cite{He_2022_CVPR} as a vision-language model (VLM) to represent diacritized Hebrew text visually.
Pretraining on images of rendered undiacritized text establishes a robust visual feature space~\cite{rust2023language} essential for modeling diacritics as visual elements.
Since undiacritized and diacritized texts share the same visual dimensions, we can further train our model on a diacritized corpus.
This allows us to train the VLM to extract distinct embeddings for each candidate~\cite{ye-etal-2021-influence} and investigate its potential to differentiate subtle diacritization patterns through visual representations.
The model's success in selecting the best candidate directly reflects its ability to learn the underlying patterns and logic of diacritization, aligning with the principles of zero-shot learning where the model generalizes to unseen classes based on learned representations.

While our system does not yet surpass Nakdimon's reported word-level accuracy of 89.75\% in all scenarios, initial results demonstrate the promise of our approach.
In an \say{oracle} setting, where the correct diacritization pattern is guaranteed to be within the model's candidate set, our model achieves word-level accuracy of 92.68\%, highlighting its robust ability to assign high scores to correct candidates when they are present.
In a more confined setting using a KNN-based candidate generator operating by character sequence alignment, the system reaches 87.87\% accuracy.
Although slightly below current benchmarks, this still reflects competitive performance.
This sets the bar for visual models performing diacritization at the word level operating in a zero-shot learning framework, selecting from full-form candidates rather than predicting individual diacritics.
These results suggest that further improvements in candidate generation and training methodologies can significantly enhance its practical effectiveness.
Furthermore, \divrit{} is the first visual model to perform a candidate-ranking task since these models have been introduced for NLP~\cite{salesky-etal-2021-robust}.
We expect its reliance on visual representations to add diacritics to outperform previous methods especially in the environments identified as strong suits of PIXEL models, such as data noised by typographical errors and adversarial character swapping, raw visual inputs that avoid OCR pipelines, and more.
Our code is publicly available online.\footnote{\url{github.com/elboheryair/DiVRit}}

\section{Related Work}
\label{sec:related_work}
Hebrew diacritization has evolved significantly with the advent of neural-based models.
Nakdan, current state-of-the-art neural network system from Dicta~\cite{shmidman-etal-2020-nakdan}, implements a hybrid approach, combining deep-learning modules with manually constructed tables, linguistic knowledge, and dictionaries.
This knowledge-intensive strategy demonstrates the effectiveness of leveraging linguistic expertise.
However, the system's dependency on language-specific rules and resources comes at the expense of its generalizability, flexibility, and high investment in linguistic engineering.

Nakdimon~\cite{gershuni-pinter-2022-restoring} represents a shift towards a purely data-driven approach.
It utilizes a streamlined character-level Bi-LSTM network, trained directly on a diacritized Hebrew corpus, an approach also shown effective for morphologically similar languages such as Arabic~\cite{belinkov-glass-2015-arabic}.
Predicting diacritics at the character level, the system achieved competitive results, with significantly reduced complexity and enhanced generalizability.
The results suggest that even without explicit linguistic knowledge, a well-trained deep learning model can effectively learn diacritization patterns, paving the way for further pure deep learning solutions.

MenakBERT~\cite{cohen-etal-2024-menakbert} advances this direction by employing a character-level transformer-based model, pretrained on a large Hebrew corpus and finetuned on diacritized texts.
In contrast to Nakdimon, which trains directly on diacritized data and achieves similar results with both LSTM and transformer-based models trained from scratch~\cite{klyshinsky-etal-2021-comparison}, MenakBERT's pretraining strategy, leveraging contextual embeddings and transfer learning, yields superior performance.

Our method is general and flexible, being fully independent of language-specific resources and applicable in diverse linguistic settings.
Additionally, inspired by ZSL principles~\cite{NIPS2013_2d6cc4b2}, for each word, it generates a relatively small set of diacritization candidates specific to that word.
While this strategy is similar in principle to Nakdan, we opt for a completely data-driven approach for candidate generation.

Zero-shot learning (ZSL) has played a significant role in the advancement of NLP, with large-scale language models like GPT-3 demonstrating notable performance across diverse tasks without task-specific training~\cite{NEURIPS2020_1457c0d6}.
Specifically, zero-shot classification leverages the representations learned by pretrained models to classify categories not encountered during training~\cite{yin-etal-2019-benchmarking}.
Works like CLIP~\cite{pmlr-v139-radford21a} and ALIGN~\cite{NEURIPS2021_50525975} have demonstrated the effectiveness of using vision and language models for ZSL classification.
ViTs extend the transformer architecture to computer vision, offering an alternative to conventional convolutional approaches~\cite{He_2022_CVPR} and demonstrating effectiveness across a range of vision tasks, showcasing their versatility~\cite{10230477,10943551,ren2025irrelevant}.
Beyond core vision tasks, ViTs have been applied to visually-grounded NLP problems~\cite{borenstein-etal-2023-phd,tai-etal-2024-pixar}, demonstrating robustness in handling dialects and non-standard language forms~\cite{munoz-ortiz-etal-2025-evaluating}.

Adhering to the zero-shot paradigm for diacritization, we use a ViT to process the visual input of the candidates.
For our VLM, we pretrained and employed a Hebrew PIXEL-based model~\cite{rust2023language}, which utilizes the Vision Transformer Masked Autoencoder (ViT-MAE) architecture~\cite{He_2022_CVPR}.
PIXEL's ability to learn robust text representations directly from raw images has been demonstrated across diverse languages and scripts.
Like in the original PIXEL architecture, our \modelname{} component consists of a 12-layer ViT encoder and an 8-layer transformer decoder~\cite{vaswani-etal-2017-attention,dosovitskiy2021imageworth16x16words}, pretrained using a masked image modeling objective analogous to BERT~\cite{devlin-etal-2019-bert}.


\section{\divrit{}}
\label{sec:methodology}
We present \divrit{}, a representation, model architecture and training protocol designed to leverage contrastive learning principles for Hebrew diacritization.
\divrit{} learns a shared representation space for the undiacritized context input and the diacritized candidates, enabling direct scoring and selection of the most contextually appropriate diacritic sequence.

\subsection{Architecture}
\label{sec:architecture}

Our processing pipeline integrates a visual candidate encoder, based on the PIXEL-base model~\cite{rust2023language}, and a context encoder (see \autoref{fig:architecture}).
This design allows us to process diacritization candidates, which are initially generated for the input undiacritized word~(\S\ref{sec:candidates_generation}).
Each of these candidates is then rendered as an image and processed by the candidate encoder, generating candidate-specific embeddings that are subsequently used for scoring.
Scoring is computed by comparing these candidate embeddings against a contextual embedding of the undiacritized word, extracted by the context encoder which may have visual and/or textual components.
Since the encoders may produce multiple embeddings for a single word, either due to their window size limitations in the visual modality or due to subword tokenization in textual encoders, we aggregate such cases into a single vector per item.
This enables simple score calculation as an inner product between the embedding of the context word and each candidate, a method well-suited for measuring alignment in high-dimensional spaces.
In all cases, we apply mean pooling to ensure a unified representation, preserving information from the entire input span~\cite{reimers-gurevych-2019-sentence}.
To enable direct comparison, both candidate and contextual embeddings are projected to a shared embedding space of identical dimensionality.
In our experiments, we explored various architectural configurations to encourage the model to develop a more nuanced and effective representation and prediction space of diacritics.

\subsection{Training the Hebrew PIXEL Model}
\label{sec:Training_the_Hebrew_PIXEL_Model}

To generate visual embeddings for the candidates, we pretrained a Hebrew PIXEL-based language model using a masked image modeling objective, as in the original PIXEL.
This pretraining was conducted on a corpus consisting of approximately 1.9GB of text from the Hebrew Wikipedia dataset and 9.8GB from the Hebrew section of the OSCAR dataset~\cite{ortiz-suarez-etal-2019-oscar}.
We filtered out examples with fewer than 30 characters, to minimize the impact of potentially noisy or incomplete text.
The model was trained for 2M steps with a total batch size of 128 across four 48GB Nvidia RTX6000 GPUs, ensuring a comparable scale of pretraining to the original PIXEL-base model.
Due to the right-to-left nature of the Hebrew script, all text images were horizontally mirrored at the full instance level.
This practice maintains script orientation alongside numerical sequences and interspersed left-to-right text. \todot{yes?}

By operating on images, PIXEL naturally handles out-of-vocabulary words (OOVs) and bypasses the need for explicit vocabulary allocation.
This vocabulary-free nature is particularly advantageous in the context of diacritization, as creating a vocabulary of diacritized tokens would likely result in an unwieldy and semantically weak set of short, less-meaningful sequences.
Unlike such token-based approaches, PIXEL learns diacritics directly from visual patterns in the images, preserving word integrity and potentially allowing the model to learn diacritic patterns as cohesive units.
To further boost the model's ability to learn diacritization patterns at the visual level, we conducted an additional pretraining phase on images of diacritized Hebrew text.
This phase continued the masked image modeling objective, but we reduced the masking ratio from 0.25 to 0.1 for patch sequence completion.
This adjustment~\cite{ye-etal-2021-influence} aims to simplify Hebrew character prediction, intending to improve the model's capacity for restoring diacritics.
The data for this phase was taken by the training dataset curated by \citet{gershuni-pinter-2022-restoring}, comprising approximately 3.4M tokens of originally-diacritized Hebrew text sourced from modern Hebrew and pre-modern Hebrew, as well as automatically-diacritized texts collected without manual validation.
This data was also used for supervised training of the diacritization task itself.

\subsection{Candidate Generation}
\label{sec:candidates_generation}

\begin{figure}[t]
    \centering
    \includegraphics[width=\linewidth]{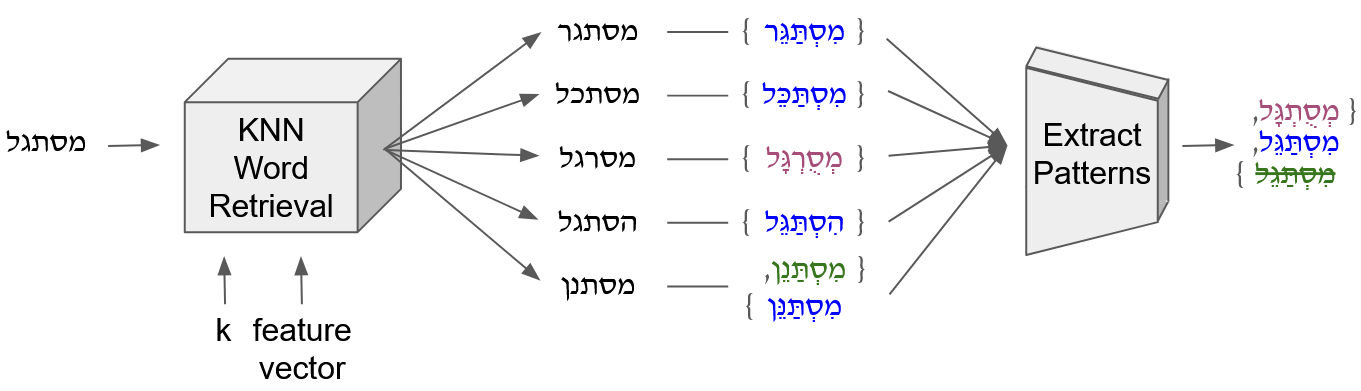}
    \caption{KNN-based candidate generator.
     The figure illustrates an example of the algorithm on a specific OOV word ($k=5, c=2$).
     Similar diacritization patterns are colored with the same color, and the output candidate set shows each pattern applied to the input word.
     The green pattern is then removed to yield a $c$-sized candidate set.}
    \label{fig:knn_algorithm}
\end{figure}

While the other diacritizers utilize model embeddings to directly predict or refine diacritization, our approach distinguishes itself by incorporating a constrained candidate generation mechanism.
Nakdimon and MenakBERT employ a minimalist, neural-only character-by-character prediction approach that allows for the generation of any potential sequence.
In contrast, Dicta's Nakdan operates as a hybrid system that leverages human-curated linguistic resources and a formal lexicon.
Before ranking, it applies morphological analysis and predicts POS tags to filter irrelevant options from its internal dictionary of pre-annotated word forms.
While this ensures high grammatical accuracy, it relies heavily on predefined rules and curated resources.
Our system, \divrit{}, bridges these two worlds.
Similar to Nakdan, we first generate candidates using a corpus-driven mechanism, capturing the empirical distribution of diacritized forms in real-world usage.
However, unlike the rule-based constraints of Nakdan’s filtering, our architecture allows the VLM to learn unconstrained representations of these candidates.
By processing candidates as images, \divrit{} captures diacritic information natively, echoing the flexible, data-driven nature of Nakdimon and MenakBERT.

To generate a set of candidates, we employ a straightforward and efficient component based on the $k$-nearest neighbors (KNN) algorithm (\autoref{fig:knn_algorithm}).
This algorithm identifies a specified number of similar words from the training data and then compiles a candidate set whose size is capped at a maximum value.
The KNN algorithm is configured with a parameter $k$ to determine the number of similar words considered, and $c$ which defines the maximal size of the returned candidate set.
The process involves an initial step of mapping each undiacritized form present in the diacritized corpus to a frequency-sorted list of its observed diacritization patterns.
Given an undiacritized input word $w$, the generator uses KNN to find the top \textit{k} most similar words (in their undiacritized form) in the corpus that are of the same length $|w|$.
Similarity is computed based on character-level matching at fixed positions, following established string similarity techniques that account for both edit distance and positional alignment~\cite{levenshtein1966binary,4160958} and leveraging the root-template morphological characteristics of Semitic languages.
Next, the generation algorithm combines the frequency-sorted pattern lists of these neighbors, preserving the order.
Subsequently, the generator applies each of these diacritical patterns to the input word while removing repetitions, creating a set of candidates, and returns the top $c$ candidates.

\begin{figure}[t]
    \centering
    \includegraphics[width=\linewidth]{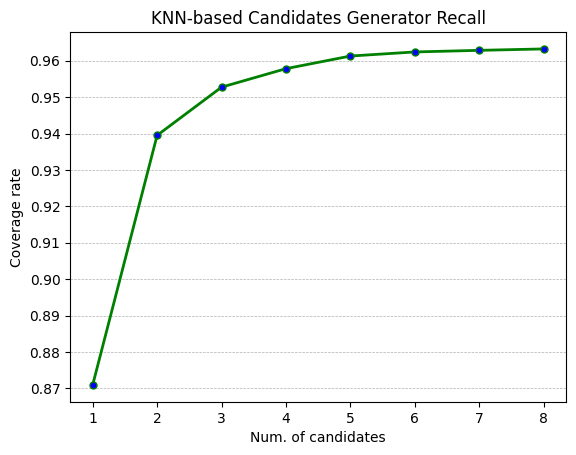}
    \caption{Coverage rate per candidate set size, \emph{i.e.}~the fraction of times the correct candidate appears in the KNN-generated set of each size.}
    \label{fig:knn_recall}
\end{figure}

Our candidate generator serves a dual purpose, acting both as a training aid and as a practical component in real-world applications.
To evaluate its potential in the latter setting, we investigated the influence of candidate set size on performance.
The coverage rate, also referred to as recall, is defined as the proportion of test words for which the correct diacritization appears among the candidates generated by our KNN-based method.
As expected, larger candidate sets improve coverage but also increase the potential for error.
\autoref{fig:knn_recall} illustrates this trade-off, exhibiting an elbow-shaped curve that suggests a candidate set size between three and four would strike a favorable balance.

\section{Experiments}
\label{sec:experiments}
We evaluate our system's performance by comparing it against Dicta, Nakdimon, and MenakBERT.
Given our data-driven methodology, the comparisons with Nakdimon and MenakBERT, which also employ data-driven approaches, are particularly relevant and provide insight for understanding our system's strengths and weaknesses.

\begin{table*}[t]
    \centering
    \begin{tabular}{lcccc}
        \toprule
        System                     & DEC             & CHA             & WOR             & VOC             \\
        \midrule
        \textsc{Majority Baseline}          & 93.79           & 90.01           & 84.87           & 86.19           \\
        \textsc{KNN Baseline}               & 96.20           & 94.09           & 87.09           & 87.39           \\
        \midrule
        \textsc{Nakdimon}                   & 97.91           & 96.37           & 89.75           & 91.64           \\
        \textsc{MenakBERT}                  & 98.82           & 97.95           & 94.12           & 95.22           \\
        \textsc{\divrit{}} (Oracle)    & 98.36           & 97.42           & 92.68           & 94.69           \\
        \textsc{\divrit{}} (KNN-based) & 96.85           & 95.03           & 87.87           & 90.38           \\
        \midrule
        \textsc{Dicta}                      & \textbf{98.94}  & \textbf{98.23}  & \textbf{95.83}  & \textbf{95.93}  \\
        \bottomrule
    \end{tabular}
    \caption{\centering
    Performance comparison of Hebrew diacritization systems.
    Results are presented on the test set in three categories: baselines, purely data-driven systems, and systems powered by manually-curated linguistic knowledge. \divrit{} variants operate over two candidates.
    All results in all tables are over a single run.
    }
    \label{tab:comparison_results}
\end{table*}

We assess our system over a test set comprised of 20K tokens collected from various Modern Hebrew sources~\cite{gershuni-pinter-2022-restoring} using four standard metrics: word-level accuracy (WOR), character-level accuracy (CHA), diacritic-level decision accuracy (DEC) and word-level vocalization preservation (VOC).
DEC evaluates the correctness of individual diacritical decisions, which may be more fine-grained than the character level: in addition to vowel marks, most characters may also carry a central dot (\emph{dagesh}) that has morphological and phonetic implications, and one letter must be marked by a distinguishing dot to denote a \emph{shin} pronunciation from a \emph{sin}.
VOC measures word-level vocalization preservation, remaining agnostic to several redundancies in vowel representations for modern pronunciation (e.g.,~the \emph{a} sound is marked by three distinct vowel diacritics).

In addition to comparison against existing diacritization packages, we implement two baselines: the first is a majority class prediction baseline, determined by the most frequent diacritization pattern for each undiacritized word in the training dataset, equivocating if it does not appear there (which counts as an incorrect prediction).
The second baseline is obtained by using the KNN-based candidate generation to produce a single-candidate set per input word ($k=1$).
This serves as a majority baseline with robustness to OOV words by assigning the most frequent diacritization of the closest in-vocabulary word.
Despite being more sophisticated, this KNN-based baseline uses no parameterized learning and no context, providing a better reference point for interpreting the effectiveness of the different models' learning and contextual understanding.

\subsection{Experimental Setup}
\label{sec:experimental_setup}

We evaluate \divrit{}'s performance across several distinct experimental setups, each designed to probe different architectural choices and training methodologies.
Each of these setups leverages a pretrained candidate encoder alongside a context encoder to generate diacritized output.
The candidate encoder, based on the \textsc{PIXEL-base} model~\cite{rust2023language}, is responsible for processing each diacritization candidate.
The context encoder provides a contextual embedding of the undiacritized word.
To determine the optimal configuration for these components, we conducted several preliminary evaluations.

To explore the role of the context encoder in our architecture, we tested two encoder variants: first, we used an additional instance of the Hebrew PIXEL model as the context encoder, creating a fully vision-based architecture.
In this setup it is possible to uniquely align the representation space across both candidate and context inputs, mapping them into a common embedding space, a concept used in models like CLIP.
This design entirely bypasses reliance on subword tokenization, a process known to be ill-fitted for nonconcatenative languages like Hebrew~\cite{gazit-etal-2025-splintering}.
In an initial evaluation of this setup, where the candidate encoder was trained directly on the classification task without prior diacritized pretraining, we observed no discernible learning.
This was evidenced by the model's stagnant 50\% accuracy in two-candidate selection and near-identical scoring of pairs, strongly suggesting a failure to attend to diacritic patterns.
Furthermore, even after optimizing the candidate encoder's pretraining (as detailed below), performance with the fully vision-based context encoder plateaued below 80\% word-level accuracy over two candidates.

\begin{table*}[t]
    \centering
    \begin{tabular}{lccccc}
        \toprule
        Experiment Setup    & FT Steps (K) & DEC             & CHA             & WOR             & VOC             \\
        \midrule
        Basic                       & 140  & 97.91           & 96.68           & 90.54           & 93.74           \\
        Extended Finetuning         & 240  & \textbf{98.17}  & \textbf{97.08}  & \textbf{91.45}  & 94.38           \\
        Auxiliary Bag-of-diacritics & 140  & 98.16           & 97.05           & 91.41           & \textbf{94.44}  \\
        Auxiliary Pos. Encoding     & 140  & 97.62           & 96.23           & 89.17           & 92.67           \\
        Balanced Data               & 140  & 97.30           & 95.65           & 86.00           & 91.78           \\
        RTL images + Extended Finetuning   & 240  & 97.59           & 96.15           & 88.60           & 92.30           \\
        \bottomrule
    \end{tabular}
    \caption{\centering
    Performance across various experimental setups, evaluated on the \emph{development set} for two-candidate selection.
    }
    \label{tab:experiments_results_two_cands}
\end{table*}

\begin{table*}[t]
    \centering
    \begin{tabular}{lccccc}
        \toprule
        Experiment Setup    & FT Steps (K) & DEC             & CHA             & WOR             & VOC             \\
        \midrule
        Basic                       & 140  & 94.24           & 90.79           & 73.55           & 82.19           \\
        Extended Finetuning         & 240  & 94.48           & 91.21           & 74.16           & 82.51           \\
        Auxiliary Bag-of-diacritics & 140  & 94.15           & 90.82           & 71.49           & 79.49           \\
        Auxiliary Pos. Encoding     & 140  & 94.74           & 91.68           & 76.56           & 83.24           \\
        Balanced Data               & 140  & 93.41           & 89.60           & 67.25           & 78.10           \\
        RTL Images + Extended Finetuning & 240  & \textbf{96.64}  & \textbf{94.67}  & \textbf{84.93}  & \textbf{89.39}  \\
        \bottomrule
    \end{tabular}
    \caption{\centering
    Three-candidate selection results. 
    }
    \label{tab:experiments_results_three_cands}
\end{table*}

To overcome these limitations in both candidate and context encoding, we developed a second variant of the model, where the context encoder is instantiated by a text-based \hebrewbert{} Hebrew language model~\cite{gueta2023largepretrainedmodelsextralarge}, and the visual candidate encoder is employed following additional pre-training.
We found that pretraining the candidate encoder on unmasking diacritized images for an additional 200K steps yielded the best performance gain, presumably because it helps the model learn diacritic patterns and establish more robust diacritic representations. 
The 200K-step pretraining proved optimal, as increasing it to 400K steps provided a slight degradation of 0.3 percentage points in word-level accuracy.
Consequently, for all our main experiments, we chose to utilize this 200K-step pretraining for the candidate encoder, and the context encoder was instantiated by the \hebrewbert{} model, which demonstrated superior contextual understanding compared to the vision-based alternative.
Following these architectural and pretraining selections, the full \divrit{} architecture was trained for an additional 240K steps on the classification objective using multiclass cross-entropy loss, with each step batching 32 examples across two GPUs.

To address potential imbalances in the frequency of different words in the training data, the sampling of undiacritized words for training iterations was based on a normalized ratio determined by the function $f(\freq) = \freq^{0.75}$, where $\freq$ represents the frequency of that word in the training data, and the exponent value is inspired by seminal findings regarding context sampling~\cite{NIPS2013_9aa42b31,levy-etal-2015-improving}.
All model hyperparameter values and training regimes were taken as the defaults from the cited models.
For the PIXEL model, We used the PangoCairo renderer, and the \emph{Noto Sans Hebrew} font from the \emph{Google Noto Sans fonts} font collection.
We pre-processed the data to be chunked into sentences following a `.' character or after the first line break following 200 characters.

\todot{Consider adding the following, maybe some or all in appendix: projection head depth/width and embedding dimensionality, embedding normalization and scoring temperature, rendering parameters (font(s), size(s), resolution/DPI, text rendering library, image size and ViT patch size), context model tokenization and window, which components are frozen vs. finetuned and with what hyperparameters, optimizer, learning rate schedule, weight decay, warmup, gradient clipping, batch sizes, training steps per phase, seed(s), number of runs.}

\subsection{Results}
\label{sec:results}

We evaluated our system under two primary evaluation schemes.
In the \textit{Oracle} setting, the correct diacritization was guaranteed to appear among the provided candidates, thereby isolating evaluation of the model's ability to distinguish among valid alternatives based on context from the abilities of our (fairly na{\"i}ve) candidate generator, which was used to fill the candidate set to size $k$.
In contrast, in the \textit{KNN-based} setting candidates were generated using the nearest-neighbor search described in the previous section, without ensuring inclusion of the gold label.
This setting reflects the real-world inference scenario and tests the system's robustness to imperfect candidate generation.
Importantly, while the candidate generation algorithm proved beneficial for training, our system's modularity allows for its replacement with other candidate generation mechanisms.

\autoref{tab:comparison_results} presents the evaluation results for \divrit{}'s best setup, as detailed in Section \ref{sec:experimental_setup}, alongside those of the other methods.
\divrit{} operates well among the ranks of similarly data-driven systems.
Moreover, both evaluation variants pass both baselines on all metrics, indicating the potential of visual representations for informing Hebrew diacritics.
The \emph{Oracle} setup greatly outperforms \textsc{Nakdimon}, especially on word-level metrics, and is on par with \textsc{MenakBERT} at the character-level metrics.
Surprisingly, despite its reliance on individual character segmentation for in-word representation, \textsc{MenakBERT} excels on word-level metrics in our evaluation.\footnote{We note that we did not implement either of the two data-driven character-level prediction systems and are relying on their own reporting.}
Finally, while Dicta understandably sets the performance ceiling, it is a hybrid system that relies on extensive, manually-curated linguistic knowledge.
Our work focuses on the more directly comparable data-driven space.
In this context, \divrit{}'s results are competitive with other neural models like Nakdimon and MenakBERT, particularly in the Oracle setting.
This demonstrates that visual representations can effectively internalize complex diacritical patterns natively, offering an interesting and more flexible alternative for scenarios where expert-crafted lexicons or rules are unavailable.


\subsection{System Component Analysis}
\label{sec:ablation_studies}

To understand the contribution of each component to our best-performing setup, we conducted a series of ablation studies over the development set provided by \citet{gershuni-pinter-2022-restoring}.
These involved systematically removing or modifying key parts of the training pipeline described in Section~\ref{sec:experimental_setup}.
The results for two-candidate KNN selection are presented in \autoref{tab:experiments_results_two_cands}, and for three-candidate selection in \autoref{tab:experiments_results_three_cands}.

\paragraph{Number of Candidates}
Comparing the overall per-setup numbers in \autoref{tab:experiments_results_two_cands} with those in \autoref{tab:experiments_results_three_cands}, it is clear that in the current setup, the increase in coverage of correct diacritic combinations afforded by the extra neighbor added to the candidate set (see \autoref{fig:knn_recall}) does not offset the deficiencies in the system's scoring mechanism.
While attempts to improve candidate generation are key to \divrit{}'s performance, increasing the number of candidates appears to be detrimental or at least insufficient.

\paragraph{Finetuning Duration}
The duration of finetuning significantly impacted the model's ability to select the correct candidate, though its effect varied across different experimental configurations.
For our primary configuration, which we call \textit{Basic}, our initial finetuning of 140K steps achieved 90.54\% word-level accuracy when choosing between two candidates.
Extending this to 240K steps further improved accuracy to 91.45\% in the same two-candidate scenario.

However, even with these gains from longer finetuning, we found the model's performance remained sensitive to the total number of candidates presented.
While longer finetuning provided benefits, the improvement was modest as the number of candidates increased.
When selecting from three candidates, accuracy only improved from 73.55\% to 74.16\%, even with the extended finetuning.
This limited generalization capability suggests the model might be learning diacritization patterns in a rather loose or context-dependent way, rather than acquiring a robust understanding of the underlying diacritization principles.

\paragraph{Auxiliary Tasks}
Building upon the \textit{Basic} setup, we explored the incorporation of an auxiliary task to further enhance performance.
In this setup, our primary task remained the prediction of the correct diacritization, using a cross-entropy loss.
To improve the discriminative representation of diacritics in the output embeddings for the diacritization candidates from the VLM, we augmented our cross-entropy loss with a secondary loss term derived from an auxiliary task, namely, prediction of the bag of diacritics present in an input image.

The final loss was computed as the sum of the primary diacritization loss and a bag-of-diacritics binary prediction loss, which was normalized by the number of candidates and weighted to ensure it did not outweigh the contrastive component:

\small
\begin{equation}
\label{eq:auxiliary_loss}
\begin{split}
L(w, &C, c_{\text{gt}}) = \text{CELoss}\left(P(c | w), \text{one\_hot}(c_{\text{gt}})\right) + \\
&\frac{0.5}{N_{\text{cands}}} \sum_{i=1}^{N_{\text{cands}}} \text{BCELoss}\left(y_{\text{diac}}(c_i), y_{\text{target\_diac}}(c_i)\right),
\end{split}
\end{equation}
\normalsize

where $w$ is the input undiacritized word, $C$ is the set of $N_{\text{cands}}$ diacritized candidates for $w$, and $c_{\text{gt}}$ is the ground truth candidate.
$y_{\text{diac}}(c_i)$ represents the predicted bag of diacritics for candidate $c_i$, and $y_{\text{target\_diac}}(c_i)$ represents the target bag of diacritics for candidate $c_i$.

While the \textit{Bag-of-Diacritics} auxiliary task led to a slight improvement in two-candidate selection accuracy, from 90.54\% to 91.41\%, and faster convergence, it did not achieve the intended generalization.
This was demonstrated by a decline in performance when selecting from three candidates, where accuracy decreased from 73.55\% to 71.49\% with its introduction.

However, longer finetuning does not always guarantee improved performance, so the exact finetuning steps for each setup are detailed in the tables.
In this auxiliary task setup, extending the duration to 240K steps significantly decreased accuracy to 82.45\% for two-candidate selection and 63.19\% for three-candidate selection.
This suggests that while auxiliary tasks might guide the model towards certain beneficial representations, they can also impose constraints on the flexibility of the embedding space.
Consequently, we shifted our auxiliary task to predict both the diacritics themselves and their corresponding positions within the image embeddings, hypothesizing that this \textit{Auxiliary Pos. Encoding} setup would embed more detailed information about the diacritic patterns.
Its performance for two-candidate selection was 89.17\%, a slight decrease compared to both the \textit{Basic} setup and the \textit{Bag-of-Diacritics} auxiliary task.
However, for three-candidate selection, this setup achieved 76.56\% accuracy.
This represents a notable improvement in generalization compared to both those setups, demonstrating that a better representation of the diacritics can indeed lead to better generalization of diacritical patterns.
While the positional encoding task is more demanding, limiting absolute peak performance on two-candidate sets, the resulting structural constraint encourages better generalization.
This advantage becomes evident as the task complexity grows with the addition of a third candidate.

\paragraph{Balanced Data}
To investigate the impact of potential bias towards the most frequent diacritization, we introduced a \textit{Balanced Data} training setup as an alternative to our standard frequency-based sampling.
This setup capped the effective training frequency of the most common diacritization pattern for each word.
Specifically, its frequency was limited to be no more than the combined frequency of all its alternative diacritization patterns (i.e.,~50\% of all occurrences).
This ensures a more equitable representation of all forms during training, preventing the model from disproportionately learning the most frequent pattern and thereby allowing rarer alternatives to be adequately learned and distinguished.
However, this approach did not yield any performance improvements.
For two-candidate selection, the model achieved only 86.0\% accuracy, and for three-candidate selection, performance dropped to 67.25\% accuracy.
This degradation suggests that artificially manipulating the frequency distribution of diacritic patterns might have disrupted the model's ability to effectively leverage contextual information provided by the context encoder.
This could occur by hindering its capacity to learn the more nuanced and naturally-occurring relationships between words and their diacritizations, which are typically reflected in a natural frequency hierarchy.

\paragraph{RTL images}
Given that diacritics are often comprised of simpler, symmetric components (dots, horizontal and vertical lines) compared to the more complex shape of Hebrew letters, mirroring the candidate images in our \textit{RTL Images} setup was hypothesized to provide a superior initial visual representation for them, thereby emphasizing diacritic features to impact model performance.
while for two-candidate selection this variant achieved an accuracy of 88.60\%, for the more challenging three-candidate selection it demonstrated a noticeably improved generalization, reaching an impressive 84.93\% accuracy.
This distinctive performance can be explained by the way vision models process input.
While they generally prioritize larger visual features, the mirroring applied to the images in this setup might have enhanced the distinctiveness of the relatively simpler, symmetric diacritic components compared to the more complex Hebrew letters.
Thus, on one hand, this processing could significantly enhance the model's ability to distinguish diacritics, but on the other hand, it might simultaneously lead to a loss of holistic information about the full Hebrew words.
This suggests a nuanced trade-off between emphasizing fine-grained diacritic features and preserving comprehensive word representation within image-based encoding.

\section{Conclusion}
\label{sec:conclusion}
We presented \divrit{}, a novel approach for Hebrew diacritization that effectively combines visual language models with contextual embeddings.
Our system addresses diacritization as a candidate selection problem, where we directly score and select the correct diacritized word from a set of alternatives by analyzing combined image and contextual embeddings, unlike other approaches that rely on embeddings for subsequent decoding or sequential prediction.
This approach, the first of its kind for visual text processing models, combines information from PIXEL for image-based diacritic candidate understanding and from \textsc{AlephBertGimmel} for contextual cues, demonstrating robust performance, particularly in the \textit{Oracle} mode, as well as benefits from adding a contrastive learning objective.

Adapting the VLM to the specific task by pretraining it on diacritized text images proved beneficial, enhancing the representation of diacritics in the candidate encoder.
Through a series of detailed ablation studies, we elucidated the crucial factors contributing to effective diacritization.
We found that finetuning duration significantly impacts performance, with 240K steps proving optimal for our primary setup.
Our exploration of auxiliary tasks showed that while this technique could lead to improved generalization through more informative embeddings, it also presented challenges that hindered overall performance, underscoring the complexities of multi-task learning.

Future work will focus on developing better candidate generation algorithms, aiming to extend the model's ability to utilize its discriminative capacities.
We will also further refine the integration of visual and contextual information, building on the promising evidence from our \textit{RTL images} experiment that better generalization is indeed achievable.
To examine the utility of ViTs for diacritization more comprehensively, we also aim to apply this methodology to other diacritics-rich languages such as Arabic and Vietnamese, including the potential for sourcing diacritics from various types of data~\citep{elgamal-etal-2024-arabic} and incorporating even more modalities in the system~\citep{shatnawi-etal-2024-automatic}.

\section*{Acknowledgements}
\label{sec:acknowledgements}
This research was supported by grant no. 2022215 from the United States–Israel Binational Science Foundation (BSF), Jerusalem, Israel.

\section*{Limitations}
\divrit{} is a stepping-stone on the road to truly structure-agnostic visual diacritization, but it's not quite there yet.
It still relies on a data-driven candidate generation method, and the best context encoder formation we found is still text-based.
Being developed over the Hebrew diacritization task, application to other languages may prove nontrivial, for example given Arabic's multiple representations for individual characters.
This is not an issue for Unicode-based encodings (which treat all forms of a character as the same) and is not a serious issue for visual Hebrew, where only five letters exhibit a mild form of this behavior.
Finally, Our experiments were conducted on a modest GPU setup, so we doubled the training steps to approximate the original PIXEL regimen.
With more resources, a larger architecture, longer pre-training, and more extensive hyperparameter tuning could likely yield better results.
In particular, \citet{gueta2023explicitmorphologicalknowledgeimproves} show that explicit morphological signals during pre-training substantially improve Hebrew performance on semantic and morphological tasks, implying that hyperparameters such as patch size and masking ratio, which determine how much morphological information must be inferred from partial image patches, may have a particularly strong influence in morphologically rich languages like Hebrew and would benefit from careful optimization.
The text rendering strategy is likewise critical~\cite{lotz-etal-2023-text}, acting as a visual tokenization step whose impact is especially pronounced in Hebrew due to the diversity of root inflections~\cite{gazit-etal-2025-splintering}.
Our current multi-phase curriculum starts with undiacritized text before the diacritized phase.
Extending pre-training on the undiacritized base and refining visual hyperparameters could lead to further gains.

\bibliography{anthology,custom}

\begin{thebibliography}{40}
\providecommand{\natexlab}[1]{#1}

\bibitem[{Belinkov and Glass(2015)}]{belinkov-glass-2015-arabic}
Yonatan Belinkov and James Glass. 2015.
\newblock \href {https://doi.org/10.18653/v1/D15-1274} {{A}rabic diacritization with recurrent neural networks}.
\newblock In \emph{Proceedings of the 2015 Conference on Empirical Methods in Natural Language Processing}, pages 2281--2285, Lisbon, Portugal. Association for Computational Linguistics.

\bibitem[{Borenstein et~al.(2023)Borenstein, Rust, Elliott, and Augenstein}]{borenstein-etal-2023-phd}
Nadav Borenstein, Phillip Rust, Desmond Elliott, and Isabelle Augenstein. 2023.
\newblock \href {https://doi.org/10.18653/v1/2023.emnlp-main.7} {{PHD}: Pixel-based language modeling of historical documents}.
\newblock In \emph{Proceedings of the 2023 Conference on Empirical Methods in Natural Language Processing}, pages 87--107, Singapore. Association for Computational Linguistics.

\bibitem[{Brown et~al.(2020)Brown, Mann, Ryder, Subbiah, Kaplan, Dhariwal, Neelakantan, Shyam, Sastry, Askell, Agarwal, Herbert-Voss, Krueger, Henighan, Child, Ramesh, Ziegler, Wu, Winter, Hesse, Chen, Sigler, Litwin, Gray, Chess, Clark, Berner, McCandlish, Radford, Sutskever, and Amodei}]{NEURIPS2020_1457c0d6}
Tom Brown, Benjamin Mann, Nick Ryder, Melanie Subbiah, Jared~D Kaplan, Prafulla Dhariwal, Arvind Neelakantan, Pranav Shyam, Girish Sastry, Amanda Askell, Sandhini Agarwal, Ariel Herbert-Voss, Gretchen Krueger, Tom Henighan, Rewon Child, Aditya Ramesh, Daniel Ziegler, Jeffrey Wu, Clemens Winter, and 12 others. 2020.
\newblock \href {https://proceedings.neurips.cc/paper_files/paper/2020/file/1457c0d6bfcb4967418bfb8ac142f64a-Paper.pdf} {Language models are few-shot learners}.
\newblock In \emph{Advances in Neural Information Processing Systems}, volume~33, pages 1877--1901. Curran Associates, Inc.

\bibitem[{Chen et~al.(2024)Chen, Adebara, and Abdul-Mageed}]{chen-etal-2024-interplay}
Wei-Rui Chen, Ife Adebara, and Muhammad Abdul-Mageed. 2024.
\newblock \href {https://doi.org/10.18653/v1/2024.naacl-long.420} {Interplay of machine translation, diacritics, and diacritization}.
\newblock In \emph{Proceedings of the 2024 Conference of the North American Chapter of the Association for Computational Linguistics: Human Language Technologies (Volume 1: Long Papers)}, pages 7559--7601, Mexico City, Mexico. Association for Computational Linguistics.

\bibitem[{Cohen et~al.(2024)Cohen, Gidron, and Pinto}]{cohen-etal-2024-menakbert}
Ido Cohen, Jacob Gidron, and Idan Pinto. 2024.
\newblock \href {https://arxiv.org/abs/2410.02417} {Menakbert -- hebrew diacriticizer}.
\newblock \emph{arXiv preprint arXiv:2410.02417}.

\bibitem[{Devlin et~al.(2019)Devlin, Chang, Lee, and Toutanova}]{devlin-etal-2019-bert}
Jacob Devlin, Ming-Wei Chang, Kenton Lee, and Kristina Toutanova. 2019.
\newblock \href {https://doi.org/10.18653/v1/N19-1423} {{BERT}: Pre-training of deep bidirectional transformers for language understanding}.
\newblock In \emph{Proceedings of the 2019 Conference of the North {A}merican Chapter of the Association for Computational Linguistics: Human Language Technologies, Volume 1 (Long and Short Papers)}, pages 4171--4186, Minneapolis, Minnesota. Association for Computational Linguistics.

\bibitem[{Dosovitskiy et~al.(2021)Dosovitskiy, Beyer, Kolesnikov, Weissenborn, Zhai, Unterthiner, Dehghani, Minderer, Heigold, Gelly, Uszkoreit, and Houlsby}]{dosovitskiy2021imageworth16x16words}
Alexey Dosovitskiy, Lucas Beyer, Alexander Kolesnikov, Dirk Weissenborn, Xiaohua Zhai, Thomas Unterthiner, Mostafa Dehghani, Matthias Minderer, Georg Heigold, Sylvain Gelly, Jakob Uszkoreit, and Neil Houlsby. 2021.
\newblock \href {https://arxiv.org/abs/2010.11929} {An image is worth 16x16 words: Transformers for image recognition at scale}.
\newblock \emph{Preprint}, arXiv:2010.11929.

\bibitem[{Elgamal et~al.(2024)Elgamal, Obeid, Kabbani, Inoue, and Habash}]{elgamal-etal-2024-arabic}
Salman Elgamal, Ossama Obeid, Mhd Kabbani, Go~Inoue, and Nizar Habash. 2024.
\newblock \href {https://doi.org/10.18653/v1/2024.acl-long.792} {{A}rabic diacritics in the wild: Exploiting opportunities for improved diacritization}.
\newblock In \emph{Proceedings of the 62nd Annual Meeting of the Association for Computational Linguistics (Volume 1: Long Papers)}, pages 14815--14829, Bangkok, Thailand. Association for Computational Linguistics.

\bibitem[{Gazit et~al.(2025)Gazit, Shmidman, Shmidman, and Pinter}]{gazit-etal-2025-splintering}
Bar Gazit, Shaltiel Shmidman, Avi Shmidman, and Yuval Pinter. 2025.
\newblock \href {https://doi.org/10.18653/v1/2025.findings-acl.1151} {Splintering nonconcatenative languages for better tokenization}.
\newblock In \emph{Findings of the Association for Computational Linguistics: ACL 2025}, pages 22405--22417, Vienna, Austria. Association for Computational Linguistics.

\bibitem[{Gershuni and Pinter(2022)}]{gershuni-pinter-2022-restoring}
Elazar Gershuni and Yuval Pinter. 2022.
\newblock \href {https://doi.org/10.18653/v1/2022.findings-naacl.75} {Restoring {H}ebrew diacritics without a dictionary}.
\newblock In \emph{Findings of the Association for Computational Linguistics: NAACL 2022}, pages 1010--1018, Seattle, United States. Association for Computational Linguistics.

\bibitem[{Gorman and Pinter(2025)}]{gorman-pinter-2025-dont}
Kyle Gorman and Yuval Pinter. 2025.
\newblock \href {https://aclanthology.org/2025.naacl-short.25/} {Don`t touch my diacritics}.
\newblock In \emph{Proceedings of the 2025 Conference of the Nations of the Americas Chapter of the Association for Computational Linguistics: Human Language Technologies (Volume 2: Short Papers)}, pages 285--291, Albuquerque, New Mexico. Association for Computational Linguistics.

\bibitem[{Gueta et~al.(2023{\natexlab{a}})Gueta, Goldman, and Tsarfaty}]{gueta2023explicitmorphologicalknowledgeimproves}
Eylon Gueta, Omer Goldman, and Reut Tsarfaty. 2023{\natexlab{a}}.
\newblock \href {https://arxiv.org/abs/2311.00658} {Explicit morphological knowledge improves pre-training of language models for hebrew}.
\newblock \emph{Preprint}, arXiv:2311.00658.

\bibitem[{Gueta et~al.(2023{\natexlab{b}})Gueta, Shmidman, Shmidman, Shmidman, Guedalia, Koppel, Bareket, Seker, and Tsarfaty}]{gueta2023largepretrainedmodelsextralarge}
Eylon Gueta, Avi Shmidman, Shaltiel Shmidman, Cheyn~Shmuel Shmidman, Joshua Guedalia, Moshe Koppel, Dan Bareket, Amit Seker, and Reut Tsarfaty. 2023{\natexlab{b}}.
\newblock \href {https://arxiv.org/abs/2211.15199} {Large pre-trained models with extra-large vocabularies: A contrastive analysis of hebrew bert models and a new one to outperform them all}.
\newblock \emph{Preprint}, arXiv:2211.15199.

\bibitem[{He et~al.(2022)He, Chen, Xie, Li, Doll\'ar, and Girshick}]{He_2022_CVPR}
Kaiming He, Xinlei Chen, Saining Xie, Yanghao Li, Piotr Doll\'ar, and Ross Girshick. 2022.
\newblock Masked autoencoders are scalable vision learners.
\newblock In \emph{Proceedings of the IEEE/CVF Conference on Computer Vision and Pattern Recognition (CVPR)}, pages 16000--16009.

\bibitem[{Klyshinsky et~al.(2021)Klyshinsky, Karpik, and Bondarenko}]{klyshinsky-etal-2021-comparison}
Edward Klyshinsky, Olga Karpik, and Alexander Bondarenko. 2021.
\newblock \href {https://doi.org/10.1007/978-3-030-71214-3_20} {A comparison of neural networks architectures for diacritics restoration}.
\newblock In \emph{Recent Trends in Analysis of Images, Social Networks and Texts}, pages 235--245. Springer.

\bibitem[{Levenshtein(1966)}]{levenshtein1966binary}
Vladimir~I. Levenshtein. 1966.
\newblock Binary codes capable of correcting deletions, insertions, and reversals.
\newblock \emph{Soviet Physics Doklady}, 10(8):707--710.

\bibitem[{Levy et~al.(2015)Levy, Goldberg, and Dagan}]{levy-etal-2015-improving}
Omer Levy, Yoav Goldberg, and Ido Dagan. 2015.
\newblock \href {https://doi.org/10.1162/tacl_a_00134} {Improving distributional similarity with lessons learned from word embeddings}.
\newblock \emph{Transactions of the Association for Computational Linguistics}, 3:211--225.

\bibitem[{Li et~al.(2021)Li, Selvaraju, Gotmare, Joty, Xiong, and Hoi}]{NEURIPS2021_50525975}
Junnan Li, Ramprasaath Selvaraju, Akhilesh Gotmare, Shafiq Joty, Caiming Xiong, and Steven Chu~Hong Hoi. 2021.
\newblock \href {https://proceedings.neurips.cc/paper_files/paper/2021/file/505259756244493872b7709a8a01b536-Paper.pdf} {Align before fuse: Vision and language representation learning with momentum distillation}.
\newblock In \emph{Advances in Neural Information Processing Systems}, volume~34, pages 9694--9705. Curran Associates, Inc.

\bibitem[{Lotz et~al.(2023)Lotz, Salesky, Rust, and Elliott}]{lotz-etal-2023-text}
Jonas Lotz, Elizabeth Salesky, Phillip Rust, and Desmond Elliott. 2023.
\newblock \href {https://doi.org/10.18653/v1/2023.emnlp-main.628} {Text rendering strategies for pixel language models}.
\newblock In \emph{Proceedings of the 2023 Conference on Empirical Methods in Natural Language Processing}, pages 10155--10172, Singapore. Association for Computational Linguistics.

\bibitem[{Mikolov et~al.(2013)Mikolov, Sutskever, Chen, Corrado, and Dean}]{NIPS2013_9aa42b31}
Tomas Mikolov, Ilya Sutskever, Kai Chen, Greg~S Corrado, and Jeff Dean. 2013.
\newblock \href {https://proceedings.neurips.cc/paper_files/paper/2013/file/9aa42b31882ec039965f3c4923ce901b-Paper.pdf} {Distributed representations of words and phrases and their compositionality}.
\newblock In \emph{Advances in Neural Information Processing Systems}, volume~26. Curran Associates, Inc.

\bibitem[{Mu{\~n}oz-Ortiz et~al.(2025)Mu{\~n}oz-Ortiz, Blaschke, and Plank}]{munoz-ortiz-etal-2025-evaluating}
Alberto Mu{\~n}oz-Ortiz, Verena Blaschke, and Barbara Plank. 2025.
\newblock \href {https://aclanthology.org/2025.coling-main.427/} {Evaluating pixel language models on non-standardized languages}.
\newblock In \emph{Proceedings of the 31st International Conference on Computational Linguistics}, pages 6412--6419, Abu Dhabi, UAE. Association for Computational Linguistics.

\bibitem[{Náplava et~al.(2021)Náplava, Straka, and Straková}]{N_plava_2021}
Jakub Náplava, Milan Straka, and Jana Straková. 2021.
\newblock \href {https://doi.org/10.14712/00326585.013} {Diacritics restoration using bert with analysis on czech language}.
\newblock \emph{Prague Bulletin of Mathematical Linguistics}, 116(1):27–42.

\bibitem[{Ortiz~Su{\'a}rez et~al.(2019)Ortiz~Su{\'a}rez, Sagot, and Romary}]{ortiz-suarez-etal-2019-oscar}
Pedro~Javier Ortiz~Su{\'a}rez, Benoit Sagot, and Laurent Romary. 2019.
\newblock \href {https://doi.org/10.14618/ids-pub-9021} {Asynchronous pipelines for processing huge corpora on medium to low resource infrastructures}.
\newblock In \emph{Proceedings of the Workshop on Challenges in the Management of Large Corpora (CMLC-7)}, pages 9--16, Cardiff, UK. Leibniz-Institut f{\"u}r Deutsche Sprache.

\bibitem[{Radford et~al.(2021)Radford, Kim, Hallacy, Ramesh, Goh, Agarwal, Sastry, Askell, Mishkin, Clark, Krueger, and Sutskever}]{pmlr-v139-radford21a}
Alec Radford, Jong~Wook Kim, Chris Hallacy, Aditya Ramesh, Gabriel Goh, Sandhini Agarwal, Girish Sastry, Amanda Askell, Pamela Mishkin, Jack Clark, Gretchen Krueger, and Ilya Sutskever. 2021.
\newblock \href {https://proceedings.mlr.press/v139/radford21a.html} {Learning transferable visual models from natural language supervision}.
\newblock In \emph{Proceedings of the 38th International Conference on Machine Learning}, volume 139 of \emph{Proceedings of Machine Learning Research}, pages 8748--8763. PMLR.

\bibitem[{Reimers and Gurevych(2019)}]{reimers-gurevych-2019-sentence}
Nils Reimers and Iryna Gurevych. 2019.
\newblock \href {https://doi.org/10.18653/v1/D19-1410} {Sentence-{BERT}: Sentence embeddings using {S}iamese {BERT}-networks}.
\newblock In \emph{Proceedings of the 2019 Conference on Empirical Methods in Natural Language Processing and the 9th International Joint Conference on Natural Language Processing (EMNLP-IJCNLP)}, pages 3982--3992, Hong Kong, China. Association for Computational Linguistics.

\bibitem[{Ren and Wang(2025)}]{ren2025irrelevant}
Qiang Ren and Junli Wang. 2025.
\newblock Irrelevant patch-masked autoencoders for enhancing vision transformers under limited data.
\newblock \emph{Knowledge-Based Systems}, 310:112936.

\bibitem[{Rust et~al.(2023)Rust, Lotz, Bugliarello, Salesky, de~Lhoneux, and Elliott}]{rust2023language}
Phillip Rust, Jonas~F. Lotz, Emanuele Bugliarello, Elizabeth Salesky, Miryam de~Lhoneux, and Desmond Elliott. 2023.
\newblock \href {https://openreview.net/forum?id=FkSp8VW8RjH} {Language modelling with pixels}.
\newblock In \emph{The Eleventh International Conference on Learning Representations}.

\bibitem[{Salesky et~al.(2021)Salesky, Etter, and Post}]{salesky-etal-2021-robust}
Elizabeth Salesky, David Etter, and Matt Post. 2021.
\newblock \href {https://doi.org/10.18653/v1/2021.emnlp-main.576} {Robust open-vocabulary translation from visual text representations}.
\newblock In \emph{Proceedings of the 2021 Conference on Empirical Methods in Natural Language Processing}, pages 7235--7252, Online and Punta Cana, Dominican Republic. Association for Computational Linguistics.

\bibitem[{Shatnawi et~al.(2024)Shatnawi, Alqahtani, and Aldarmaki}]{shatnawi-etal-2024-automatic}
Sara Shatnawi, Sawsan Alqahtani, and Hanan Aldarmaki. 2024.
\newblock \href {https://doi.org/10.18653/v1/2024.naacl-long.233} {Automatic restoration of diacritics for speech data sets}.
\newblock In \emph{Proceedings of the 2024 Conference of the North American Chapter of the Association for Computational Linguistics: Human Language Technologies (Volume 1: Long Papers)}, pages 4166--4176, Mexico City, Mexico. Association for Computational Linguistics.

\bibitem[{Shmidman et~al.(2020)Shmidman, Shmidman, Koppel, and Goldberg}]{shmidman-etal-2020-nakdan}
Avi Shmidman, Shaltiel Shmidman, Moshe Koppel, and Yoav Goldberg. 2020.
\newblock \href {https://doi.org/10.18653/v1/2020.acl-demos.23} {{N}akdan: Professional {H}ebrew diacritizer}.
\newblock In \emph{Proceedings of the 58th Annual Meeting of the Association for Computational Linguistics: System Demonstrations}, pages 197--203, Online. Association for Computational Linguistics.

\bibitem[{Sick et~al.(2025)Sick, Engel, Hermosilla, and Ropinski}]{10943551}
Leon Sick, Dominik Engel, Pedro Hermosilla, and Timo Ropinski. 2025.
\newblock \href {https://doi.org/10.1109/WACV61041.2025.00091} {Attention-guided masked autoencoders for learning image representations}.
\newblock In \emph{2025 IEEE/CVF Winter Conference on Applications of Computer Vision (WACV)}, pages 836--846.

\bibitem[{Socher et~al.(2013)Socher, Ganjoo, Manning, and Ng}]{NIPS2013_2d6cc4b2}
Richard Socher, Milind Ganjoo, Christopher~D Manning, and Andrew Ng. 2013.
\newblock \href {https://proceedings.neurips.cc/paper_files/paper/2013/file/2d6cc4b2d139a53512fb8cbb3086ae2e-Paper.pdf} {Zero-shot learning through cross-modal transfer}.
\newblock In \emph{Advances in Neural Information Processing Systems}, volume~26. Curran Associates, Inc.

\bibitem[{Tai et~al.(2024)Tai, Liao, Suglia, and Vergari}]{tai-etal-2024-pixar}
Yintao Tai, Xiyang Liao, Alessandro Suglia, and Antonio Vergari. 2024.
\newblock \href {https://doi.org/10.18653/v1/2024.findings-acl.874} {{PIXAR}: Auto-regressive language modeling in pixel space}.
\newblock In \emph{Findings of the Association for Computational Linguistics: ACL 2024}, pages 14673--14695, Bangkok, Thailand. Association for Computational Linguistics.

\bibitem[{Tsarfaty et~al.(2019)Tsarfaty, Sadde, Klein, and Seker}]{tsarfaty-etal-2019-whats}
Reut Tsarfaty, Shoval Sadde, Stav Klein, and Amit Seker. 2019.
\newblock \href {https://doi.org/10.18653/v1/D19-3044} {What`s wrong with {H}ebrew {NLP}? and how to make it right}.
\newblock In \emph{Proceedings of the 2019 Conference on Empirical Methods in Natural Language Processing and the 9th International Joint Conference on Natural Language Processing (EMNLP-IJCNLP): System Demonstrations}, pages 259--264, Hong Kong, China. Association for Computational Linguistics.

\bibitem[{Vaswani et~al.(2017)Vaswani, Shazeer, Parmar, Uszoreit, Jones, Gomez, Kaiser, and Polosukhin}]{vaswani-etal-2017-attention}
Ashish Vaswani, Noam Shazeer, Niki Parmar, Jakob Uszoreit, Llion Jones, Aidan~N. Gomez, Lukasz Kaiser, and Illia Polosukhin. 2017.
\newblock \href {https://proceedings.neurips.cc/paper_files/paper/2017/file/3f5ee243547dee91fbd053c1c4a845aa-Paper.pdf} {Attention is all you need}.
\newblock In \emph{Advances in Neural Information Processing Systems}, volume~30.

\bibitem[{Xian et~al.(2019)Xian, Lampert, Schiele, and Akata}]{8413121}
Yongqin Xian, Christoph~H. Lampert, Bernt Schiele, and Zeynep Akata. 2019.
\newblock \href {https://doi.org/10.1109/TPAMI.2018.2857768} {Zero-shot learning—a comprehensive evaluation of the good, the bad and the ugly}.
\newblock \emph{IEEE Transactions on Pattern Analysis and Machine Intelligence}, 41(9):2251--2265.

\bibitem[{Ye et~al.(2021)Ye, Li, Wang, Bolte, Ma, Yih, Ren, and Khabsa}]{ye-etal-2021-influence}
Qinyuan Ye, Belinda~Z. Li, Sinong Wang, Benjamin Bolte, Hao Ma, Wen-tau Yih, Xiang Ren, and Madian Khabsa. 2021.
\newblock \href {https://doi.org/10.18653/v1/2021.emnlp-main.573} {On the influence of masking policies in intermediate pre-training}.
\newblock In \emph{Proceedings of the 2021 Conference on Empirical Methods in Natural Language Processing}, pages 7190--7202, Online and Punta Cana, Dominican Republic. Association for Computational Linguistics.

\bibitem[{Yin et~al.(2019)Yin, Hay, and Roth}]{yin-etal-2019-benchmarking}
Wenpeng Yin, Jamaal Hay, and Dan Roth. 2019.
\newblock \href {https://doi.org/10.18653/v1/D19-1404} {Benchmarking zero-shot text classification: Datasets, evaluation and entailment approach}.
\newblock In \emph{Proceedings of the 2019 Conference on Empirical Methods in Natural Language Processing and the 9th International Joint Conference on Natural Language Processing (EMNLP-IJCNLP)}, pages 3914--3923, Hong Kong, China. Association for Computational Linguistics.

\bibitem[{Yujian and Bo(2007)}]{4160958}
Li~Yujian and Liu Bo. 2007.
\newblock \href {https://doi.org/10.1109/TPAMI.2007.1078} {A normalized levenshtein distance metric}.
\newblock \emph{IEEE Transactions on Pattern Analysis and Machine Intelligence}, 29(6):1091--1095.

\bibitem[{Zhou et~al.(2023)Zhou, Liu, Bae, He, Samaras, and Prasanna}]{10230477}
Lei Zhou, Huidong Liu, Joseph Bae, Junjun He, Dimitris Samaras, and Prateek Prasanna. 2023.
\newblock \href {https://doi.org/10.1109/ISBI53787.2023.10230477} {Self pre-training with masked autoencoders for medical image classification and segmentation}.
\newblock In \emph{2023 IEEE 20th International Symposium on Biomedical Imaging (ISBI)}, pages 1--6.

\end{thebibliography}






\end{document}